\documentclass[11pt]{article}

\usepackage[final]{acl}

\usepackage{times}
\usepackage{latexsym}

\usepackage[T1]{fontenc}

\usepackage[utf8]{inputenc}

\usepackage{microtype}

\usepackage{inconsolata}

\usepackage{graphicx}
\usepackage{url}
\usepackage{latexsym}
\usepackage{booktabs}
\usepackage{graphicx}
\usepackage{multirow}
\usepackage{booktabs}
\usepackage{tabularx}
\usepackage{booktabs}
\usepackage{float}
\usepackage{xcolor}
\usepackage[most]{tcolorbox}
\tcbuselibrary{listings, skins, breakable}
\tcbuselibrary{documentation}
\usepackage{fvextra}
\usepackage[svgnames]{xcolor}
\usepackage{ragged2e}
\definecolor{lightgreen}{RGB}{0, 100, 0}
\definecolor{navyblue}{RGB}{0, 51, 102}
\usepackage{enumitem}

\title{Iterative Critique-Refine Framework for Enhancing LLM Personalization}

\author{ Durga Prasad Maram\textsuperscript{1}, Dhruvin Gandhi\textsuperscript{1}, Zonghai Yao\textsuperscript{1}, Gayathri Akkinapalli\textsuperscript{1}, \\ {\bf Franck Dernoncourt\textsuperscript{2},} {\bf Yu Wang\textsuperscript{3},} {\bf Ryan A. Rossi\textsuperscript{2},} {\bf Nesreen K. Ahmed\textsuperscript{4}} \\ \textsuperscript{1}University of Massachusetts Amherst, \textsuperscript{2}Adobe Research, \textsuperscript{3}University of Georgia, \\ \textsuperscript{4}Cisco AI Research \\ \small{\textbf{Correspondence:} \href{mailto:dmaram@umass.edu}{dmaram@umass.edu}} }

\newcommand{\sysname}{PerFine}
\begin{document}
\maketitle
\begin{abstract}
Personalized text generation requires models not only to produce coherent text but also to align with a target user’s style, tone, and topical focus. Existing retrieval-augmented approaches such as LaMP and PGraphRAG enrich profiles with user and neighbor histories, but they stop at generation and often yield outputs that drift in tone, topic, or style. We present \textbf{\sysname}, a unified, training-free critique–refine framework that enhances personalization through iterative, profile-grounded feedback. In each iteration, an LLM generator produces a draft conditioned on the retrieved profile, and a critic LLM - also conditioned on the same profile - provides structured feedback on tone, vocabulary, sentence structure, and topicality. The generator then revises, while a novel knockout strategy retains the stronger draft across iterations. We further study additional inference-time strategies such as Best-of-$N$ and Topic Extraction to balance quality and efficiency. Across Yelp, Goodreads, and Amazon datasets, PerFine consistently improves personalization over PGraphRAG, with G-Eval gains of +$8.5$–$12$\%, steady improvements over $3$–$5$ refinement iterations, and scalability with increasing critic size. These results highlight that post-hoc, profile-aware feedback offers a powerful paradigm for personalized LLM generation that is both training-free and model-agnostic.
\end{abstract}

\section{Introduction}

\begin{figure}[!ht]
  \includegraphics[width=\columnwidth]{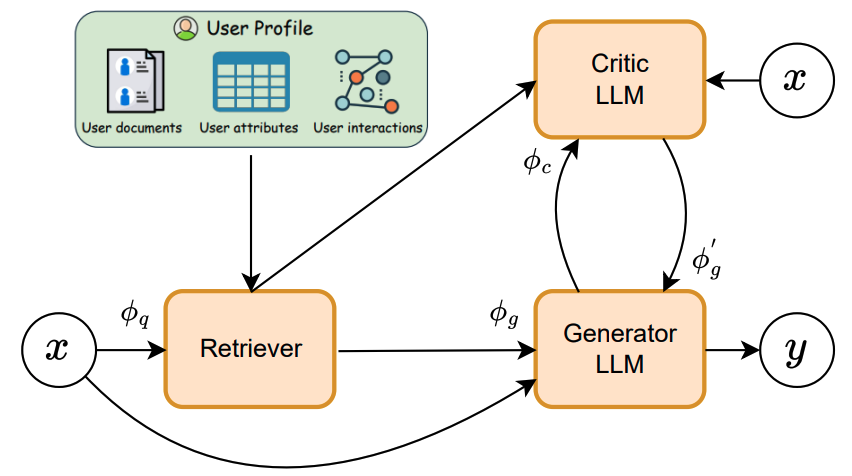}
  \caption{Overview of our framework for personalized text generation. User profile information is retrieved to guide the generator, whose outputs are iteratively critiqued and refined by PerFine, enabling multi-round personalization.}
  \label{fig:overview}
\end{figure}

Personalization is increasingly important for real-world applications such as HCI, recommender systems, conversational assistants, and natural language generation \cite{recsurvey,alhafni-etal-2024-personalized}. Prior work on personalized text generation has largely relied on retrieval-augmented generation (RAG) from user profiles. However, the challenge extends beyond retrieving relevant history: generated text must also match a user’s \emph{style}, tone, and topical focus. Benchmarks such as LaMP, LongLaMP, and PGraphRAG broaden the evaluation of personalization in long-text generation \cite{Salemi2023LaMPWL,Au2025PersonalizedGR,Kumar2024LongLaMPAB}. More recently, graph-based retrieval has been proposed to address the cold-start problem where users have sparse profiles, by leveraging neighbor user profiles in user-centric knowledge graphs \cite{Au2025PersonalizedGR}. Yet existing personalization methods primarily optimize what to retrieve and how many few-shot samples to include. They often lack a \emph{post-hoc, profile-aware refinement} step that aligns both style and content once the full draft has been generated. We conjecture that such a refinement stage, enabled by a \emph{critic LLM}, is crucial: it allows evaluation of the entire output against the user profile, correction of tone drift and topical gaps that retrieval cannot address, and avoidance of local heuristics from stepwise scoring. A profile-aware critic also acts as a model-agnostic oracle at inference, enabling easy integration with existing generation pipelines without costly retraining. By decoupling retrieval from alignment and enabling feedback to progressively reshape outputs, iterative refinement offers a practical approach to deployable personalized systems.

Personalized text generation is particularly challenging \cite{cogwriter,personalizationsurvey}: it requires capturing nuanced writing styles and implicit preferences, retrieving relevant content, often from noisy and heterogeneous profiles, which poses practical challenges, and tailoring that content to align with those preferences. A natural way to address this complexity is through \textbf{iterative refinement}, inspired by how humans revise their writing based on feedback. Automated feedback has proven effective in addressing LLM errors such as hallucinations, unfaithful reasoning, and biased content \cite{feedbacksurvey}, either during training, during generation, or through post-hoc correction. Post-hoc correction is especially suitable for personalization, since it evaluates the \emph{entire} draft after generation and provides comprehensive feedback beyond stepwise scoring. Building on this paradigm, we introduce an external iterative refinement framework that generates structured feedback on \textbf{style} and \textbf{content}—the two core dimensions of personalized text generation.

We propose \textbf{\sysname{}}, a training-free iterative refinement framework that operates entirely at inference time. Figure~\ref{fig:overview} illustrates the architecture. We first retrieve top-$k$ entries from a user–item graph using GraphRAG. A generator LLM produces an initial draft $y_0$. A critic LLM, conditioned on the same profile, returns feedback along four dimensions: tone consistency, vocabulary match, sentence structure, and topic relevance. The generator then revises the draft accordingly based on the feedback. The loop treats style and content as first-class constraints and is compatible with any base model that accepts natural language feedback.

Beyond this basic \emph{\sysname{}} loop, we explore inference-time extensions that trace the quality–efficiency frontier (Figures~\ref{fig:buget-performance-critic}, \ref{fig:buget-performance-gen}). The \emph{Knockout} strategy retains the stronger draft across iterations, while \emph{Topic Extraction} distills the profile into compact style and content hints. \emph{Best-of-$N$}, layered on Knockout, samples multiple revisions per iteration and allows the critic to select the best candidate, yielding the highest quality at higher token cost. Balancing quality and efficiency, \emph{\sysname{} with Knockout} is our default.

We address three questions. First, does a profile-grounded critique–refine loop improve personalized text generation compared to state-of-the-art methods such as LaMP and PGraphRAG? Second, how do critic scale and inference strategy trade off quality against efficiency? Third, how do improvements evolve across refinement iterations, and when do they plateau?

Our results show consistent gains. On G-Eval, \sysname{} outperforms PGraphRAG by +9.48\% (Yelp), +8.80\% (Goodreads), and +11.62\% (Amazon). METEOR also improves across datasets (0.180$\to$0.194 on Yelp, 0.206$\to$0.214 on Goodreads, 0.190$\to$0.202 on Amazon). Gains accumulate over 3–5 refinement rounds before leveling off. Larger critics provide monotonic improvements (e.g., Yelp G-Eval climbs from 0.544 to 0.577 when scaling from 14B to 32B). Among inference-time strategies, Best-of-$N$ variant yields the strongest quality, while Knockout variant balances performance and efficiency. On the email-writing task, \sysname{} also outperforms LaMP on the CMCC dataset with a G-Eval gain of +4.4\%, demonstrating cross-task generalization beyond review writing.

\noindent\textbf{Contributions.}
\begin{itemize} [noitemsep, topsep=2pt]
    \item We frame personalized text generation as \emph{profile-grounded, post-hoc critique and refinement}, separating retrieval from alignment.
    \item We introduce \textbf{\sysname{}}, a training-free iterative refinement framework for personalized text generation.
    \item We propose inference-time strategies (Knockout, Topic Extraction, Knockout + Best-of-$N$) that enable flexible trade-offs between quality and efficiency.
    \item We provide an empirical study across 3 real-world datasets, showing stable gains within a few iterations, monotonic benefits from larger critics, and practical trade-offs that keep the method training-free and model-agnostic.
    \item We also perform profile-referenced evaluation alongside ground-truth comparison, across multiple LLM judges.
\end{itemize}

\section{Related Work}

\textbf{Personalization:} RAG-based approaches are prevalent in personalized text generation \cite{Salemi2023LaMPWL,graphrag}, as they avoid reliance on LLMs' long-context processing capability. More recent works include an EM–reinforced self-training approach that teaches LLMs to reason over a user’s history and writing style \cite{Salemi2025ReasoningEnhancedSF}, and RL-based methods that optimize retrievers using LLM feedback \cite{salemi2024optimizationmethodspersonalizinglarge}. Personalized LLM alignment with multiple user preferences is introduced using a novel multi-objective RL paradigm in \cite{personalizedsoups}. Our approach is tangential to these, relying on inference-time compute scaling, leveraging natural language feedback in a RAG setting.
\\
\textbf{Feedback for LLMs:} Post-hoc feedback methods generally operate in a self-refinement setting or with an external model, which may also be trained, using either scalar scores or natural language as feedback  \cite{feedbacksurvey}. Self-Refine introduced a framework where a single powerful LLM generates, critiques, and iteratively improves its own outputs \cite{selfrefine}. Meanwhile, Reflexion \cite{reflexion} stores previously generated feedback and outputs to enhance the next iteration of feedback. External critic models \cite{selfedit,algo} offer the flexibility to decouple the feedback process from the main LLM doing the target task, and also allow working with closed-source models like ChatGPT. Trained critic models that enable smaller LLMs to generate targeted feedback include RL-based approaches \cite{rl4f} and also methods that provide feedback on intermediate steps \cite{refiner}.

However, most of these approaches focus on coding, math, and QA benchmarks, and do not address personalized text generation. In contrast, PerFine targets profile-grounded personalization through (1) a \textbf{profile-conditioned critic} that generates user-specific feedback, (2) \textbf{structured personalization dimensions} capturing attributes such as tone, vocabulary, and topicality, and (3) \textbf{an inference-time, model-agnostic refinement framework} that avoids retraining or fine-tuning. These personalization-specific components distinguish PerFine from existing self-reflection methods and address challenges unique to personalized generation.

\section{Proposed Framework}
\paragraph{Problem Definition:}
Given an input $x$ provided by a user $u$, and a target output generation $y$, the goal of the \emph{personalized text generation} problem is to generate $\tilde{y}$ that is aligned to the style and content of $y$, conditioned on the user's personal data $P_u$ (e.g., documents, histories, preferences). This is achieved by transforming the input $x$ into a personalized input $\tilde{x}$ using query ($\phi_q$) and generation ($\phi_g$) functions, along with a retriever module $R$ with an optional parameter $k$ , before passing it to a text generation module, an LLM denoted as $\textrm{Generator}$.
\begin{align*}
\tilde{x} &= \phi_g\big(x, R(\phi_q(x), P_u, k)\big) \\
      \tilde{y}  &= \textrm{Generator}(\tilde{x})
\end{align*}
\paragraph{\sysname{}:} Our refinement framework, as depicted in Figure \ref{fig:overview}, iteratively critiques and refines outputs to produce the final personalized generation $\tilde y$. The framework comprises the following core components: (1) Retriever, (2) Generator, and (3) Critic. The Retriever fetches the subset of user's profile data most relevant to user query. The Generator produces the initial draft and refines the outputs based on feedback in subsequent rounds, while the Critic generates feedback. Both the Generator and Critic components are LLMs and are conditioned on the user profile.
\paragraph{Retriever:} Since a user’s profile can be large, using it in its entirety may lead to high computational costs, over-reliance on an LLM’s ability to process long contexts, and the inclusion of irrelevant profile information. To mitigate these issues, we leverage the retrieval-augmented generation (RAG) paradigm \cite{rag}, which uses only the most relevant user-profile entries, controlled by the hyperparameter $k$, as the context.
\[
\mathcal{R}(P_u) = R(\phi_q(x), P_u, k)
\]
where $R$ is the retriever and $\mathcal{R}(P_u)$ represents the top-$k$ subset of the user profile $P_u$ ($\mathcal{R}(P_u) \subseteq P_u$). The query function $\phi_q(x)$ is an identity function in our setting.
\paragraph{Generator:} The generator LLM produces an initial draft \(y_0\) from the user query and the top-\(k\) relevant profile entries, using the generation prompt construction function \(\phi_g\). It also refines the output in further rounds based on the personalized feedback given by critic using the refinement prompt construction function $\phi_g^{'}$.
\[
\begin{gathered}
y_0 = \textrm{Generator}\big(\phi_g(x, \mathcal{R}(P_u))\big) \\
y_{t+1} = \textrm{Generator}\big(\phi_g'(x, y_t, f_t, \mathcal{R}(P_u))\big), \\
0 \le t < T
\end{gathered}
\]
where $y_t$ and \(f_t\) are the output and feedback in the $t^{th}$ iteration, while $T$ is the maximum feedback iteration count.
\paragraph{Critic:}
The Critic LLM, conditioned on $x$ and $\mathcal{R}(P_u)$ is instructed to provide feedback $f_t$ on $y_t$ along the following four personalization criteria:
\begin{itemize}[noitemsep, topsep=2pt]
    \item \textbf{Tone Consistency}: Evaluate whether the tone and sentiment align with the user's writing style.
    \item \textbf{Vocabulary Match}: Evaluate whether the vocabulary level is consistent with the user’s lexicon.
    \item \textbf{Sentence Structure}: Evaluate if the sentence lengths, complexity, and syntactic structures align with that of the user's style.
    \item \textbf{Topic Relevance}:  Evaluate if the generated content is relevant to the query, free of off-topic information, and inclusive of important aspects.
\end{itemize}
\[
f_t = \textrm{Critic}\big(\phi_c(x, y_t,\mathcal{R}(P_u) )\big) , 0\le t \le T
\]
$\phi_c$ is the feedback prompt construction function.
\paragraph{\sysname{} Setting:} The generator and critic LLMs operate in a zero-shot setting. The refinement process stops when a predefined stopping criterion is met, which here is a fixed number of iterations $T$.\\
In our setup, a user’s profile, consisting of text samples from both the user and their interacting neighbors, is represented as a user-centric bipartite graph, with users and items as node partitions and text samples connecting them as edges. We employ GraphRAG, where retrieval is performed over the graph to extract most relevant profile entries. Formally, for any user u, we define the user profile $P_u$
as the set of previous texts written by user u, and the set of texts written by other users v (interacting neighbors) for the same items connected to user u.
\[
\begin{gathered}
P_u = \{(u, j) \in E\} \; \cup \; \{(v, j) \in E \mid (u, j) \in E\}, \\
\forall j \in I, \;\; u, v \in U, \;\; u \neq v
\end{gathered}
\]
where $U$ is the set of user nodes, $I$ is the set of item nodes and $E$ is the set of edges \cite{graphrag}.
The prompt construction functions $\phi_g$, $\phi_c$, $\phi_g^{'}$,  and the query $x$ template are shared in Appendix \ref{subsec:prompts}.
\section*{Inference-time strategies}
We explore three additional inference-time variants built on top of \sysname{} to improve its performance.
\\ \\
\textbf{(1) \sysname{} + Knockout}: In this setting, after each generation, the critic LLM compares the output $y_t$ at iteration $t$ with that of the previous round $y_{t-1}$, and selects the more personalized output, which then proceeds to the next round of feedback and refinement. To determine which output is more personalized, critic is instructed to evaluate the alignment in terms of style and topical relevance, referencing the profile $\mathcal{R}(P_u)$ and the query $x$.
\[
y_t = \textrm{Critic}\big(\phi_k(x, y_t,y_{t-1}, \mathcal{R}(P_u))\big), t> 0
\]
where $\phi_k$ is the knockout prompt construction function (shared in Figure \ref{fig:knockout_prompt}). $y_t$ is followed by $f_t$ and $y_{t+1}$. For the default setting, we operate the critic in this Best-of-$T$ kind of framework, which retains the most personalized generation across iterations. The results are shown in Table
\ref{tab:main_table}.
\\ \\
\textbf{(2) PerFine + Knockout + Best-of-$\textbf{N}$}: In this setting, for every refinement step, we do Best-of-$N$ sampling and ask the critic to compare the $n$ sampled revisions and select the more personalized output, which is then passed on to the next round of knockout, feedback, and refinement steps.
\[
\begin{gathered}
y_{t+1}^1,..., y_{t+1}^n
    = \textrm{Generator}\big(\phi_g'(x, y_t, f_t, \mathcal{R}(P_u))\big) \\
y_{t+1}
    = \textrm{Critic}\big(\phi_n(x, y_{t+1}^1,..., y_{t+1}^n, \mathcal{R}(P_u))\big)
\end{gathered}
\]
where $\phi_n$ is the Best-of-$N$ prompt construction function (shared in Figure \ref{fig:best_of_n_prompt}). $y_{t+1}$ is followed by knockout, feedback and refinement steps.
\\ \\
\textbf{(3) PerFine + Topic Extraction}:  In this setting, instead of conditioning the critic on raw text samples from the user profile, we extract personalized aspects along the dimensions of style and content, which are then used as context. The writing style is derived from the user’s history, while the content aspects are taken from the neighbors' history. By leveraging an explicitly summarized representation of the personalized aspects, this approach enables the critic to reference the profile information more easily during feedback generation at each iteration, while also reducing the length of the input context.
\begin{align*}
S &= \textrm{Topic\_Extractor}\big(\phi_{ts}(\mathcal{R}(P_u)_U)\big) \\
C &= \textrm{Topic\_Extractor}\big(\phi_{tc}(\mathcal{R}(P_u)_N)\big) \\
f_t &= \textrm{Critic}\big(\phi_t(x, y_t, S, C)\big) , 0\le t \le T
\end{align*}
where $f_t$ is the feedback, $\mathcal{R}(P_u)_U$ has samples only from user's history, while $\mathcal{R}(P_u)_N$ has samples from neighbor's history. $S$ and $C$ are the extracted style and content topics by a Topic\_Extractor module, an LLM. The extraction prompts $\phi_{ts},\phi_{tc}$ are shared in Figure \ref{fig:topic_extraction_prompts}. Critic prompt $\phi_{t}$ is shared in Figure \ref{fig:critic_topic_extraction}.
\begin{table}[t]
\renewcommand{\arraystretch}{1.2}
\setlength{\tabcolsep}{4.5pt}
\centering
\small
\begin{tabular}{l|l|cc}
\hline
Dataset & Method & METEOR & G-Eval  \\
\hline
\multirow{3}{*}{Yelp}
 & LaMP        & 0.156 & 0.435      \\
 & PGraphRAG   & 0.180 & 0.496      \\
 & \begin{tabular}[c]{@{}l@{}}\sysname{}\\+ Knockout\end{tabular} & \textbf{0.194}$_{\pm 0.0003}$ & \textbf{0.543}$_{\pm 0.0008}$  \\
\hline
\multirow{3}{*}{Goodreads}
 & LaMP        & 0.193 & 0.528     \\
 & PGraphRAG   & 0.206 & 0.534    \\
 & \begin{tabular}[c]{@{}l@{}}\sysname{}\\+ Knockout\end{tabular} & \textbf{0.214}$_{\pm 0.0015}$ & \textbf{0.581}$_{\pm 0.0031}$  \\
\hline
\multirow{3}{*}{Amazon}
 & LaMP        & 0.183 & 0.491    \\
 & PGraphRAG   & 0.190 & 0.525    \\
 & \begin{tabular}[c]{@{}l@{}}\sysname{}\\+ Knockout\end{tabular} & \textbf{0.202}$_{\pm 0.0014}$ & \textbf{0.586}$_{\pm 0.0017}$  \\
\hline
\end{tabular}
\caption{\textbf{Baseline Comparison}. The table compares the performance of \textbf{\sysname{}+Knockout} (mean$_{\pm \text{std}}$) against two baselines: \textbf{LaMP}, where the user profile contains samples only from the user’s own history, and \textbf{PGraphRAG}, which also includes samples from the profile histories of neighboring users. The output of PGraphRAG is taken as the initial generation $y_0$, over which \sysname{}+Knockout performs the feedback refinement process. Qwen-2.5-14B-Instruct is used as the critic LLM, while Llama-3.1-8B-Instruct serves as the generator LLM for \sysname{} and the baseline methods.}
\label{tab:main_table}
\end{table}
\section{Experiments Design}
\paragraph{Dataset:}We evaluate our approach on the product review writing task using the AgentSociety Challenge dataset \cite{agentsociety}, which is a curated collection of user–item-review triplets (user, review, item) from Yelp, Amazon \cite{amazon}, and Goodreads \cite{goodreads1,goodreads2}. In the bipartite graph, one partition represents users, while the other represents items (businesses, products, or books being reviewed). An edge corresponds to a review.

For each dataset, we prepare development and test sets, each containing 2,500 randomly sampled users with no overlap. To construct the user–item graph, we randomly select one review per user, corresponding to a user–item pair
$(u,j)$, and assign it to the split (development/test). The remaining reviews written by user
$u$ are included as edges in the data graph. Furthermore, all reviews for item $j$ written by other users are also added as edges in the graph. We consider only users who have at least one entry in their profile history.

For Amazon reviews \cite{amazon}, we consider the domains Industrial and Scientific, Musical Instruments, and Video Games. For Goodreads reviews \cite{goodreads1,goodreads2}, we consider the domains Comics, Poetry, and Children’s Books. To describe a business in Yelp data, we consider the \textit{city}, \textit{state}, \textit{attributes}, and \textit{categories} fields. To describe Amazon products, we consider the \textit{title}, \textit{description}, and \textit{categories} fields. For Goodreads, we use the \textit{title} and \textit{description} fields from raw data. Reviews that are not in English are filtered out.
\begin{table}[t]
\renewcommand{\arraystretch}{1.2}
\centering
\small
\begin{tabular}{l|l|cc}
\hline
Dataset & Critic LLM & METEOR & G-Eval \\
\hline
\multirow{6}{*}{Yelp}
 & Llama-3.1-8B         & 0.200 & 0.521 \\
 & Qwen-2.5-7B        & 0.191 & 0.531       \\
 & Qwen-2.5-14B    & 0.195 & 0.544       \\
 & Qwen-2.5-32B         & 0.208 & 0.577 \\
 & gpt-5-mini         & \textbf{0.211} & \textbf{0.590}  \\
\hline
\multirow{6}{*}{Goodreads}
 & Llama-3.1-8B         & 0.211 & 0.558 \\
 & Qwen-2.5-7B        & 0.213 & 0.572       \\
 & Qwen-2.5-14B    & 0.216 & 0.585       \\
 & Qwen-2.5-32B         & 0.222 & 0.591 \\
 & gpt-5-mini          & \textbf{0.230}  & \textbf{0.610}  \\
\hline
\multirow{6}{*}{Amazon} & Llama-3.1-8B         & 0.203 & 0.561 \\
 & Qwen-2.5-7B        & 0.200 & 0.579       \\
 & Qwen-2.5-14B    & 0.204 & 0.588       \\
 & Qwen-2.5-32B         & 0.209 & 0.603 \\
 & gpt-5-mini         & \textbf{0.211}& \textbf{0.614} \\
\hline
\end{tabular}
\caption{Evaluation of different critic models in the \sysname{}+Knockout setup (all are instruct models). Llama-3.1-8B-Instruct is the generator.}
\label{tab:Model_ablation}
\end{table}
\paragraph{Experimental Setup:} In our main experiments, we use Qwen-2.5-14B-Instruct as the critic LLM, while Llama-3.1-8B-Instruct serves as the generator for both \sysname{} and all baseline methods.
We used vLLM\footnote{\url{https://github.com/vllm-project/vllm}} with a maximum completion token limit of 512 for generations and a temperature of 0.6 for both LLMs. The feedback iteration count $T$ is set to 5. We choose Contriever \cite{contriever} as the retriever with top-$k$ set to 4, where a max of $k$ entries are retrieved from each of the user's and neighbor's histories. The same retrieved top-$k$ are used for baselines as well. For \sysname{} + Knockout + Best-of-$N$, we choose $n$ to be 3.
\paragraph{Evaluation:}For evaluation, we use both term-based matching metrics and LLM-as-a-Judge metrics. For term-based matching, we use METEOR, while for LLM-as-a-Judge, we use G-Eval \cite{geval}, in which an LLM is prompted to generate an absolute score based on the comparison between the generation and the ground-truth reference. Gemma-3-27B-Instruct is used as the backbone. Details on G-Eval and the evaluation prompt are in Appendix \ref{subsec:evaluation_prompts}. For all ablation studies, we use the same baseline generations to ensure a fair comparison of the refinement impact across experimental settings. Ablation results under the main setting use a single run from Table \ref{tab:main_table}.

\section{Results}

\subsection{Baseline Comparison}
We compare our method with 2 personalized baselines. They are (1) LaMP \cite{Salemi2023LaMPWL}, in which the augmented information for RAG consists of only the target user's history. (2) PGraphRAG \cite{graphrag}, where the retrieval is performed to fetch information from both the target user and the neighbors (from the interaction graph). Table~\ref{tab:main_table} demonstrates that \sysname{}+Knockout consistently outperforms the baselines, achieving improvements of 9.48\% on Yelp, 8.80\% on Goodreads, and 11.62\% on Amazon in the G-Eval scores. A case study illustrating the end-to-end iterative refinement process is presented in Appendix \ref{subsec:example}.

\subsection{Ablation Studies}
\label{subsec:ablation}
\begin{figure}[H]
\includegraphics[width=\columnwidth]{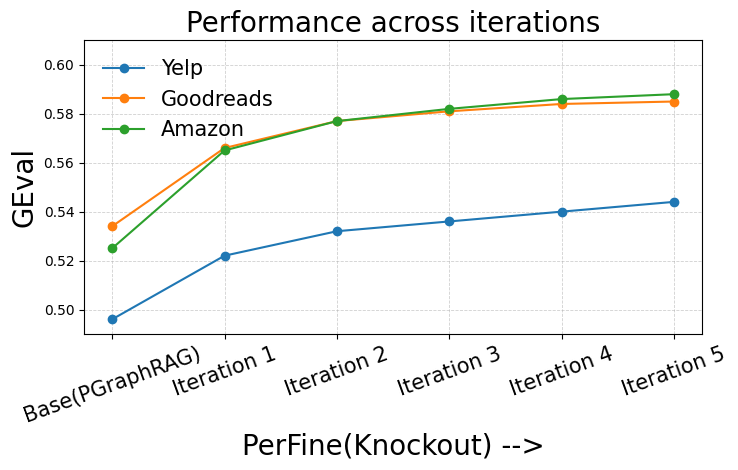}
  \caption{Performance across iterations on Yelp, Goodreads, and Amazon. \sysname{}+Knockout starts from the PGraphRAG baseline and exhibits steady improvements, with gains plateauing after a few iterations.}
  \label{fig:iterations}
\end{figure}
\paragraph{Iteration Count:} The performance trend across time is shown in Figure \ref{fig:iterations}. We observe that the scores increase with the number of iterations and converge after a few rounds of feedback, suggesting controlled and incremental alignments with the user profile over time. This evidence highlights the impact of incorporating feedback in enhancing the personalization capability of LLMs.

\paragraph{Critic Model:} In Table \ref{tab:Model_ablation}, we present an ablation study by varying both the size and the family of the critic model. All experimental configurations outperform the baselines, demonstrating that the \sysname{} framework is compatible with a wide range of base models. Evaluating Qwen-2.5 in 7B, 14B, and 32B variants, we observe that performance consistently improves with larger critics. Larger critics provide more targeted feedback and refinement suggestions, thereby simplifying the generator’s incorporation of corrections. We also evaluate a self-refinement configuration in which the critic and generator are identical. This setting is resource-efficient, as it eliminates the need for a separate critic model and reduces the overall memory footprint. Even in this setting, \sysname{} outperforms the baselines. The performance gain achieved with the closed-source model (gpt-5-mini) suggests that powerful proprietary models can be readily leveraged via APIs for generating feedback in \sysname{} framework. Furthermore, the observed gains with gpt-5-mini were obtained under a low reasoning efficiency setting, indicating that large reasoning models used as critics may be highly effective in the context of personalization and may lead to faster convergence.
\begin{figure}[!ht]
  \includegraphics[width=\columnwidth]{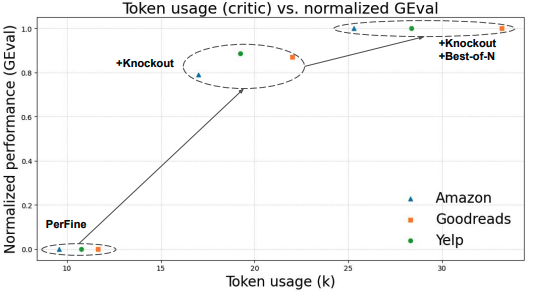}
  \caption{Visualization of the critic's token usage (prompt + completion) vs normalized G-Eval performance.
  Notably, \sysname{}+Knockout improves performance, while \sysname{}+Knockout+Best-of-N achieves the highest scores, with increased token cost. Considering both efficiency and effectiveness, we ultimately select \sysname{}+Knockout.}
  \label{fig:buget-performance-critic}
\end{figure}
\begin{figure}[!ht]
  \includegraphics[width=\columnwidth]{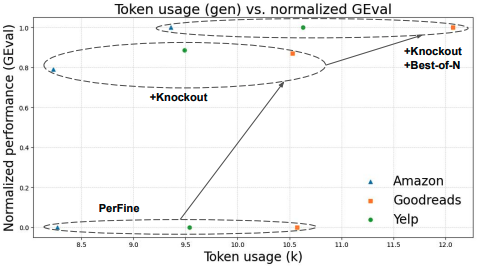}
  \caption{Figure showing generator's token usage (prompt + completion) vs normalized G-Eval performance.
  While the token usage for \sysname{} and \sysname{}+Knockout is similar, the token footprint increases for \sysname{}+Knockout+Best-of-N due to the sampling of multiple revisions, while yielding only a marginal improvement in performance.}
  \label{fig:buget-performance-gen}
\end{figure}

\begin{table}[h]
\resizebox{\linewidth}{!}{%
\renewcommand{\arraystretch}{1.2}
\centering
\small
\begin{tabular}{l|l|cccc}
\hline
Dataset & Method & METEOR & G-Eval & \# token  & \# token \\
& & & & (critic) & (gen)
\\
\hline
\multirow{3}{*}{Yelp}
 & \sysname{}        & 0.185 & 0.505 & 10.77 & 9.54      \\
 & +Knockout   & 0.195 & 0.544 & 19.23 & 9.49     \\
 &  + Best-of-N          & \textbf{0.197} & \textbf{0.549} & 28.36 & 10.63 \\
\hline
\multirow{3}{*}{Goodreads}
 & \sysname{}        & 0.209 & 0.565  & 11.65 & 10.57     \\
 & +Knockout   & 0.216 & 0.585  & 22 & 10.53   \\
 & + Best-of-N         & \textbf{0.219} & \textbf{0.588} & 33.18 & 12.07 \\
\hline
\multirow{3}{*}{Amazon}
 & \sysname{}        & 0.197 & 0.558  & 9.58 & 8.27     \\
 & + Knockout   & 0.204 & 0.588  & 17 & 8.23   \\
 & + Best-of-N        & \textbf{0.206} & \textbf{0.596} &  25.28 & 9.36 \\
\hline
\end{tabular}
}
\caption{Comparison of performance and token usage (prompt+completion) across various inference strategies. We observe that the scores improve as we scale the inference time token usage. However, the additional gains from Best-of-N sampling are marginal, while the associated token usage increases significantly. Token count is per query for 5 iterations in K. Llama-3.1-8B is the generator, and Qwen-2.5-14B is the critic.}
\label{tab:method_ablation}
\end{table}
\paragraph{Inference Strategies:}Table \ref{tab:method_ablation} compares the performance of different inference-time strategies for the critic. The total token usage per query by the critic and generator is shown in Figure~\ref{fig:buget-performance-critic} and Figure~\ref{fig:buget-performance-gen}, respectively. The \sysname{}+Knockout strategy, when combined with Best-of-$N$, outperforms all other configurations. In both \sysname{}+Knockout and \sysname{}+Knockout+Best-of-$N$ settings, multiple candidate revisions are laid out for the critic to compare, either across iterations or within the same iteration, which increases the likelihood of producing an optimized output. However, this performance gain comes at the cost of higher token usage and increased latency (Figures \ref{fig:buget-performance-critic}, \ref{fig:buget-performance-gen}). The increase in generator token usage for the \sysname{}+Knockout+Best-of-$N$ setting arises from sampling multiple revisions, while the spike in critic token usage comes from selecting the best among the $N$. The incremental improvement provided by Best-of-$N$ sampling is relatively small, and \sysname{}+Knockout offers a nice balance between performance and efficiency. The quality of \sysname{}+Topic\_Extraction setting is discussed in Appendix \ref{subsec:topic_extraction}. Per-query average wall-clock times across strategies are compared in Appendix \ref{subsec:latency}.

\paragraph{Generator Model:} In Table \ref{tab:generator_ablation}, we present an ablation study varying both the size and the family of the generator. For studying the
impact of model size, we evaluate the Ministral-3 model series in 3B, 8B, and 14B variants, and observe that \sysname{} performs better than the baseline across all experimental settings.

\begin{table}[H]
\renewcommand{\arraystretch}{1.1}
\resizebox{\linewidth}{!}{%
\centering
\begin{tabular}{l|cc|cc}
\hline
\multirow{3}{*}{Dataset}  &
\multicolumn{2}{c}{G-Eval-Style} &
\multicolumn{2}{c}{G-Eval-Content} \\
\cline{2-5}
&  \multirow{2}{*}{PGraphRAG} & PerFine & PGraphRAG & PerFine \\
& & + Knockout & &+ Knockout \\
\hline
Yelp & 0.405 &	\textbf{0.473} & 0.427 & \textbf{0.502} \\

Goodreads & 0.444 &	\textbf{0.498} & 0.461 & \textbf{0.528}\\

Amazon & 0.425	& \textbf{0.491} & 0.448 & \textbf{0.527}\\
\hline
\end{tabular}
}
\caption{
Performance of \sysname{}+Knockout across style and content dimensions. Llama-3.1-8B-Instruct is the generator and gpt-5-mini is the critic.}
\label{tab:personalization_att_ablation}
\end{table}

\paragraph{Personalization Attributes:} We modify the G-Eval prompt to compare the generated text with the ground truth separately along style and content dimensions. Table \ref{tab:personalization_att_ablation} presents the evaluation results for our best-performing setting, using gpt-5-mini as the critic, showing consistent improvements for \sysname{} across all datasets.

\begin{table}[h]
\renewcommand{\arraystretch}{1.2}
\centering
\small
\begin{tabular}{lcc}
\hline
Method & METEOR & G-Eval \\
\hline
PGraphRAG & 0.184 & 0.524 \\
\hline
\multicolumn{3}{l}{PerFine+Knockout} \\
\hline
Original prompts   & 0.200 & 0.589 \\
Critic-Variant-1   & 0.201 & 0.583 \\
Critic-Variant-2   & 0.197 & 0.584 \\
Knockout-Variant-1 & 0.201 & \textbf{0.592} \\
Knockout-Variant-2 & \textbf{0.202} & 0.589 \\
\hline
\end{tabular}
\caption{Prompt sensitivity study of \sysname{}+Knockout on a 500-sample subset of the Amazon dataset. Each variant perturbs a single prompt while keeping all other prompts unchanged.}
\label{tab:prompt_sensitivity}
\end{table}

\begin{table*}[t]
\renewcommand{\arraystretch}{1.2}
\centering
\small
\begin{tabular}{l|l|cc|cc}
\hline
\multirow{2}{*}{Dataset} & \multirow{2}{*}{Generator LLM} &
\multicolumn{2}{c}{METEOR} &
\multicolumn{2}{c}{G-Eval} \\
\cline{3-6}
& & PGraphRAG & PerFine+Knockout & PGraphRAG & PerFine+Knockout \\
\hline
\multirow{4}{*}{Yelp}
& Llama-3.1-8B-Instruct  & 0.180 & \textbf{0.195} & 0.496 & \textbf{0.544} \\
& DeepSeek-R1-Distill-Qwen-7B & 0.167 & \textbf{0.174} & 0.435 & \textbf{0.461} \\
& Ministral-3-3B-Instruct     & 0.191 & \textbf{0.194} & 0.550 & \textbf{0.567} \\
& Ministral-3-8B-Instruct     & 0.193 & \textbf{0.203} & 0.566 & \textbf{0.580} \\
& Ministral-3-14B-Instruct    & 0.200 & \textbf{0.204} & 0.581 & \textbf{0.600} \\
\hline
\multirow{4}{*}{Goodreads}
& Llama-3.1-8B-Instruct  & 0.206 & \textbf{0.216} & 0.534 & \textbf{0.585} \\
& DeepSeek-R1-Distill-Qwen-7B & 0.182 & \textbf{0.187} & 0.492 & \textbf{0.502} \\
& Ministral-3-3B-Instruct     & 0.185 & \textbf{0.193} & 0.578 & \textbf{0.594} \\
& Ministral-3-8B-Instruct    & 0.193 & \textbf{0.200} & 0.586 & \textbf{0.595} \\
& Ministral-3-14B-Instruct     & 0.191 & \textbf{0.201} & 0.590 & \textbf{0.607} \\
\hline
\multirow{4}{*}{Amazon}
& Llama-3.1-8B-Instruct  & 0.190 & \textbf{0.204} & 0.525 & \textbf{0.588} \\
& DeepSeek-R1-Distill-Qwen-7B & 0.181 & \textbf{0.185} & 0.508 & \textbf{0.522} \\
& Ministral-3-3B-Instruct      & 0.187 & \textbf{0.193} & 0.606 & \textbf{0.619} \\
& Ministral-3-8B-Instruct      & 0.192 & \textbf{0.200} & 0.616 & \textbf{0.631} \\
& Ministral-3-14B-Instruct    & 0.192 & \textbf{0.197} & 0.616 & \textbf{0.637} \\
\hline
\end{tabular}
\caption{
Comparison of different generator LLMs in the \sysname{}+Knockout setup. Qwen-2.5-14B-Instruct is the critic. \sysname{}+Knockout consistently improves performance across all generator LLMs.}
\label{tab:generator_ablation}
\end{table*}

\paragraph{Prompt Sensitivity:} To assess sensitivity to prompt changes, we perturb the critic and knockout prompts using a few variants and evaluate performance on a subset of 500 samples from the Amazon dataset. The generator prompt primarily consists of the input query and profile information, and therefore remains effectively fixed. For the critic, \textit{Critic-Variant-1} paraphrases the instructions and feedback criteria, while \textit{Critic-Variant-2} reorders and paraphrases the feedback criteria. For the knockout step, \textit{Knockout-Variant-1} paraphrases the task and criterion descriptions, while \textit{Knockout-Variant-2} reorders the comparison criteria and uses shorter wording. For each variant, all other prompts are kept unchanged. As shown in Table \ref{tab:prompt_sensitivity}, the improvements are stable, with the critic prompt exhibiting slight sensitivity to prompt variations. The standard deviation across variants remains below 1\% of the average score for both metrics, indicating robustness to minor prompt modifications.

\subsection{Profile Based Evaluation}
Since personalization requires generating text that captures complex user preference patterns, evaluating alignment with the user distribution is a stronger indicator of personalization than comparing against a single ground-truth text. Therefore, we also assess \sysname{} by measuring how well the generated output aligns with the user profile distribution as represented by the user’s profile history. Using G-Eval with the top-$k$ profile history as the sole reference, we evaluate how well the method captures user-specific patterns and stylistic nuances. With gpt-5-mini as the critic and Gemma-3-27B-Instruct as the LLM judge, the results in Table \ref{tab:profile_ref_results} show that PerFine consistently outperforms PGraphRAG, achieving improvements of 8.47\% on Yelp, 2.99\% on Goodreads, and 4.76\% on Amazon, demonstrating its effectiveness in modeling nuanced user preferences. The evaluation prompt is provided in Figure \ref{fig:geval-profile}.

\begin{table}[H]
\renewcommand{\arraystretch}{1.1}
\centering
\small
\begin{tabular}{lcc}
\hline
Dataset & PGraphRAG & PerFine+Knockout \\
\hline
Yelp         & 0.59 & \textbf{0.64} \\
Goodreads     & 0.67 & \textbf{0.69} \\
Amazon        & 0.63 & \textbf{0.66} \\
\hline
\end{tabular}
\caption{
Performance of \sysname{}+Knockout in the user profile-based evaluation setting, where top-$k$ samples ($k$=4) from the author’s writing are used as the reference text. Llama-3.1-8B-Instruct is used as the generator and gpt-5-mini as the critic.
}
\label{tab:profile_ref_results}
\end{table}

To further evaluate robustness with respect to the choice of LLM judge, we evaluated \sysname{} under the profile-referenced evaluation setting using multiple LLM judges from different model families, including Gemma-3-27B-Instruct, Qwen3-32B, and Llama-3.3-70B-Instruct. The percentage gains in G-Eval scores compared to the PGraphRAG baseline are shown in Table \ref{tab:llm_judge_results}.
\begin{table}[H]
\resizebox{\linewidth}{!}{%
\centering
\begin{tabular}{lccc}
\hline
LLM-Judge & Amazon & Yelp & Goodreads \\
\hline
Gemma-3-27B-It  & 4.76\% & 8.47\% & 2.99\% \\
Qwen3-32B       & 6.45\% & 10.00\% & 4.54\% \\
Llama-3.3-70B-It & 8.19\% & 14.00\% & 7.69\% \\
\hline
\end{tabular}
}
\caption{G-Eval percentage gains of \sysname{}+Knockout over PGraphRAG baseline under different LLM judges in the profile-referenced evaluation setting.}
\label{tab:llm_judge_results}
\end{table}

Our method consistently outperforms the baselines across all judges, demonstrating robustness to evaluator choice. While the magnitude of improvements increases with stronger judges, the relative ranking remains stable, suggesting that larger LLM judges provide more reliable and discriminative evaluation signals.

\subsection{Personalized Email Writing}
To further assess the cross-task generalization of our framework, we also evaluate it on the email-writing task, another long-form writing problem. We use the CMCC \cite{cmcc} dataset, and take the test data split from \cite{ditto}. It consists of 27 queries from 19 different authors covering six different controversial topics. We compare \sysname{}, with Qwen-2.5-32B-Instruct as the critic, against the LaMP baseline (CMCC does not have a graph structure), with a 1024-token generation limit. All prompts used for review writing are adapted accordingly for this task, without the neighbors' histories. Our approach consistently outperforms LaMP on METEOR and across multiple LLMs as judges, under both ground-truth and profile-referenced evaluation, as shown in Table \ref{tab:email_results}.

\begin{table}[t]
\renewcommand{\arraystretch}{1.2}
\setlength{\tabcolsep}{9pt}
\centering
\small
\begin{tabular}{lcc}
\hline
 & LaMP & PerFine+Knockout \\
\hline
METEOR & 0.257 & \textbf{0.265} \\
\hline
\multicolumn{3}{l}{G-Eval (Ground-truth reference)} \\
\hline
Gemma-3-27B-It   & 0.450 & \textbf{0.470} \\
Qwen3-32B        & 0.333 & \textbf{0.350} \\
Llama-3.3-70B-It & 0.362 & \textbf{0.420} \\
\hline
\multicolumn{3}{l}{G-Eval (Profile reference)} \\
\hline
Gemma-3-27B-It   & 0.691 & \textbf{0.693} \\
Qwen3-32B        & 0.597 & \textbf{0.627} \\
Llama-3.3-70B-It & 0.636 & \textbf{0.677} \\
\hline
\end{tabular}
\caption{Performance of \sysname{}+Knockout and the LaMP baseline on the CMCC email-writing dataset. G-Eval scores are reported under different LLM judges, using either the ground-truth email or the author's past emails as the reference.}
\label{tab:email_results}
\end{table}

\section{Findings}
Existing personalization methods focus on what to retrieve and how many few-shot samples to include, but lack a post-hoc, profile-aware refinement step that aligns style and content. Our iterative refinement framework achieves this effectively in a training-free, inference-time feedback loop, covering key personalization dimensions. Empirical evidence showed that personalized feedback, obtained by conditioning the critic on a relevant subset of the user profile retrieved via a retriever model, improved text generation quality. By ablating the feedback and generator models across different sizes and families, we observed that PerFine improved performance in all settings, showing that the personalized refinement process is model-agnostic. Larger critic models provided more fine-grained feedback leading to larger performance gains. The improvements were monotonic across iterations, suggesting that additional rounds can be beneficial, though gains plateau after a few iterations, allowing the stopping point to be tuned. We also proposed and evaluated multiple refinement methods to scale inference-time computation. As expected, increasing the token budget (for both critic and generator) improves performance, but there is a trade-off between cost, latency, and gains. Personalized feedback also improved both stylistic and content alignment and modeled complex user preferences more effectively, as evidenced by the profile-referenced evaluation. On the CMCC email-writing task, PerFine also outperformed LaMP, suggesting that personalized feedback is beneficial for long-form writing more broadly.

\section{Conclusion}
In this work, we introduced \sysname{}, a training-free, profile-aware critique--refine framework for personalized text generation. By separating retrieval from post-hoc alignment through structured, profile-conditioned feedback, \sysname{} improves over strong RAG baselines without user-specific fine-tuning. We design multiple inference-time scaling mechanisms with different performance–efficiency trade-offs, allowing flexible selection for deployment on cost, latency, and quality requirements.  This work opens up promising research directions in the area of feedback for personalized LLMs.

\section*{Limitations}

\looseness=-1 One of the main limitations of our approach is the fixed number of feedback iterations. Different queries require different levels of refinement, and dynamically judging when to stop helps to avoid both overcorrection and undercorrection, while also reducing token usage and latency. In the \sysname{} framework, although smaller critics improve performance, the most significant gains come from bigger LLMs. In resource-constrained deployment settings, this presents a challenge, making it crucial to train smaller LLMs as personalized critics for improved efficiency. However, obtaining training signals for such supervision is difficult, and exploring synthetic data creation techniques may be helpful. Another limitation is the sensitivity to retrieval quality, where irrelevant profile entries may propagate noisy signals to both the generator and critic leading to retrieval-induced hallucination, especially in the initial generation rounds. Several possible extensions to \sysname{} can be explored to enhance its capabilities further. One promising direction is to make the retrieval process dynamic, where the top-$k$ profile is updated at every iteration based on feedback. In-context learning can also be effective in guiding the critic to structure its feedback in a specific way tailored to the target task.

\section*{Acknowledgment}

We used ChatGPT (OpenAI) throughout the manuscript for language polishing. All AI-generated text was reviewed and edited by the authors, who verified its accuracy and take full responsibility for the content, claims, and analyses.

\bibliography{custom}
\clearpage
\appendix
\onecolumn
\section{Appendix}
\label{sec:appendix}
\subsection{Topic Extraction}
\label{subsec:topic_extraction}
\begin{table*}[h]
\centering
\begin{tabular}{l|l|ccccc}
\hline
Dataset & Method & METEOR & G-Eval & \# token  & \# token & \# token \\
& & & & (critic) & (gen) & (topics)
\\
\hline
\multirow{2}{*}{Yelp}
 & \sysname{}        & 0.185 & 0.505 & 10.77 & 9.54  & -    \\
  & \sysname{}  + {Topic\_Extraction}   & \textbf{0.188} & \textbf{0.513}  & 7.73 & 9.48 & 2.60  \\
\hline
\multirow{2}{*}{Goodreads}
 & \sysname{}        & \textbf{0.209} & \textbf{0.565}  & 11.65 & 10.57 & -    \\
  & \sysname{}  + {Topic\_Extraction}   & 0.209 & 0.557  & 7.73 & 10.48 & 2.81  \\
\hline
\multirow{2}{*}{Amazon}
 & \sysname{}         & 0.197 & 0.558  & 9.58 & 8.27  & -   \\
 & \sysname{}  + {Topic\_Extraction}   & \textbf{0.199} & \textbf{0.559}  & 7.73 & 8.19  & 2.09 \\
\hline
\end{tabular}
\caption{Performance and Token usage for Topic Extraction method. The topic extraction method outperforms the baselines and achieves performance comparable to the vanilla setting, while requiring fewer critic tokens during feedback rounds. However, it introduces an initial overhead for extracting topics, which increases the overall token consumption. Qwen-2.5-32B-Instruct is used for extracting the topics.}
\label{tab:topic_extraction_tab}
\end{table*}

\subsection{Latency Analysis}
\label{subsec:latency}
Table \ref{tab:latency} reports the per-query average wall-clock time over the 2,500 test samples of each dataset, measured on a single A100 (80GB) GPU with five refinement rounds using the vLLM framework, with Llama-3.1-8B-Instruct as the generator and Qwen-2.5-14B-Instruct as the critic. Knockout and Best-of-$N$ increase the overall latency due to the additional critic calls. The latency also varies depending on the sizes of the LLMs used and the available computational resources. Using the same LLM as both the generator and critic (a setup that also outperforms the baselines) avoids loading multiple model weights and further reduces latency.

\begin{table*}[h]
\centering
\begin{tabular}{l|ccc}
\hline
Dataset & \sysname{} & \sysname{}+Knockout & \sysname{}+Knockout+Best-of-N \\
\hline
Yelp      & 2.29 & 3.69 & 5.68 \\
Goodreads & 2.62 & 4.17 & 6.65 \\
Amazon    & 3.48 & 4.73 & 6.53 \\
\hline
\end{tabular}
\caption{Per-query average wall-clock time (in seconds) for different refinement strategies.}
\label{tab:latency}
\end{table*}

\onecolumn
\clearpage
\subsection{Prompts}
\label{subsec:prompts}
\vspace{2em}
\newcounter{boxcounter}
\setcounter{boxcounter}{4}
\refstepcounter{boxcounter}
\begin{tcolorbox}[
  width=\textwidth,
  fontupper=\small\ttfamily,
  title={Query template $x$ (Yelp)}
]
\label{fig:query}
\begin{Verbatim}[breaklines=true,breaksymbol={}]
Generate a review text written by a user who has given an overall rating of [TARGET_RATING] for a business with following details:
[BUSINESS_DETAILS]
\end{Verbatim}
\end{tcolorbox}
Figure \ref{fig:query}: Query Template

\vspace{2em}
\refstepcounter{boxcounter}
\begin{tcolorbox}[
  width=\textwidth,
  fontupper=\small\ttfamily,
  breakable,
  enhanced,
  title={Generation Prompt $\phi_g$ (Yelp)},
]
\label{fig:generation_prompt}
\begin{Verbatim}[breaklines=true,breaksymbol={}]
Given a target user’s past reviews, a business description, and reviews from other users for the same business, your task is to generate a review that the target user might write.
Below is the list of reviews from the target user on different businesses: 

#User Profile:
SAMPLE 1:
[SAMPLE 1 Text]
...

SAMPLE 2:
[SAMPLE 2 Text]
...

Consider a business with the following details: 

[BUSINESS_DETAILS]

Others have reviewed this business before. Below is a list of their reviews:

#Neighbor Profile:
SAMPLE 1:
[SAMPLE 1 Text]
...

SAMPLE 2:
[SAMPLE 2 Text]
...

Now, generate a review from the target user who has given an overall rating of [TARGET_RATING] for the business mentioned above, without any additional explanation, adhering to the following instructions. 
- Analyze the target user’s previous reviews to capture their writing style, tone, sentiment and content preferences.
- Use the business details to understand its features, benefits, and intended use.
- Consider the opinions expressed in other users’ reviews to identify common themes regarding various aspects of the business.
- Generate a review that reflects how the target user would likely perceive the business, consistent with their past preferences and reviewing habits.
Use the format: 'Review text:'.
\end{Verbatim}
\end{tcolorbox}
Figure \ref{fig:generation_prompt}: Generation Prompt Template

\onecolumn
\refstepcounter{boxcounter}
\begin{tcolorbox}[
  width=\textwidth,
  fontupper=\small\ttfamily,
  title={Critic Prompt $\phi_c$(Yelp)},
  breakable,
  enhanced,
]
\label{fig:critic_prompt}
\begin{Verbatim}[breaklines=true,breaksymbol={}]
You are a Personalization Critic Module. Specifically, you will analyze user review texts written for businesses. Your task is to evaluate how well the *Generated Output* is personalized for the Target User and to give feedback along the specified criteria.

# Your Inputs
Query - The user's request containing the business details.
Rating - the overall rating the target user has given for the business.
Generated Output - review text to be evaluated.
User Profile - past reviews written by the target user for other businesses.
Neighbor Profile - reviews from other users for this business.

# Feedback Criteria 
- Tone Consistency: Does the tone and sentiment in the **Generated Output** align with the user’s writing style and the target review rating?
- Vocabulary Match: Does the vocabulary level in the **Generated Output** align with that of the user?
- Sentence Structure: Are the sentence lengths, complexity, and syntactic structures in the **Generated Output** similar to those used by the user? 
- Topic Relevance: Is the output content relevant to the query, free of off-topic information, and inclusive of important aspects?

# Analysis Strategy
- Analyze the user's writing style from the **User Profile**. Your feedback for the first three criteria (Tone Consistency, Vocabulary Match, Sentence Structure) should solely be on style and should not include any content-related suggestions. To evaluate tone, use profile reviews with sentiments that align closely with the target rating.
- Consider opinions expressed in other users’ reviews to identify themes and aspects of the business from the **Neighbor Profile** relevant to the query. Use these insights to identify any off-topic or missing aspects in the **Generated Output**. Your feedback for the last criterion (Topic Relevance) should solely address the content and be grounded in **Neighbor Profile**.

You can output 'No further improvement needed' along a criteria if the performance along the criteria is exceptional.

# Output Format

Return your output in the following JSON format:
{
"Tone Consistency": **Feedback text**,
"Vocabulary Match": **Feedback text**,
"Sentence Structure": **Feedback text**,
"Topic Relevance": **Feedback text**
}

## Test Inputs

#Query:
Generate a review text written by a user who has given an overall rating of [TARGET_RATING] for a business with following details:
[BUSINESS_DETAILS]

#Generated Output:
[GENERATED_OUPUT]

#User Profile:
SAMPLE 1:
[SAMPLE 1 Text]
...

SAMPLE 2:
[SAMPLE 2 Text]
...

#Rating:
[TARGET_RATING]

#Neighbor Profile:
SAMPLE 1:
[SAMPLE 1 Text]
...

SAMPLE 2:
[SAMPLE 2 Text]
...

Your output should be a valid json object in ```json ``` block without any additional explanation.

\end{Verbatim}
\end{tcolorbox}
Figure \ref{fig:critic_prompt}: Critic Prompt Template

\vspace{2em}
\refstepcounter{boxcounter}
\begin{tcolorbox}[
  width=\textwidth,
  fontupper=\small\ttfamily,
  title={Refinement Prompt $\phi_g^{'}$ (Yelp)}
]
\label{fig:refinement_prompt}
\begin{Verbatim}[breaklines=true,breaksymbol={}]
[GENERATION PROMPT]

Your previously generated review is:
[GENERATED_OUTPUT]

Your review has room for improvement. The feedback on your review from an expert is:
[FEEDBACK]

Based on the improvements suggested in the feedback, please improve your review, without any additional explanation.
Use the format: 'Review text:'.
\end{Verbatim}
\end{tcolorbox}
Figure \ref{fig:refinement_prompt}: Refinement Prompt Template. The [GENERATION PROMPT] in the above template is shared in Figure \ref{fig:generation_prompt}
\clearpage
\refstepcounter{boxcounter}
\begin{tcolorbox}[
  width=\textwidth,
  fontupper=\small\ttfamily,
  title={Knockout Prompt $\phi_k$ (Yelp)}
]
\label{fig:knockout_prompt}
\begin{Verbatim}[breaklines=true,breaksymbol={}]
You are an impartial evaluator of style and content alignment. You will be evaluating a review text about a particular business written by an author. Below are samples of the author's writing containing past review texts of various businesses, review samples from other users for the business being reviewed, the input query, and two reviews.

# Author's Writing:
SAMPLE 1:
[SAMPLE 1 Text]
...

SAMPLE 2:
[SAMPLE 2 Text]
...

# Reviews from other users:
SAMPLE 1:
[SAMPLE 1 Text]
...

SAMPLE 2:
[SAMPLE 2 Text]
...

# Query:
[QUERY]

# Review A:
[GENERATED_OUTPUT_A]

# Review B:
[GENERATED_OUTPUT_B]

# Task
Analyze the samples given under *Author's Writing* to identify the author's writing style. Consider the opinions expressed in *Reviews from other users* above to understand common themes and aspects of the business. Determine which of the two reviews is more likely to have been written by the author. Consider each review's similarity with regards to (1) tone (2) vocabulary level (3) sentence structure (4) tonal alignment with the target rating (5) avoidance of off-topic information (6) inclusion of information relevant to the query based on other user reviews.

# Output Format:
You must return the winning review option along with a brief explanation for your choice.
Your output should be a valid json object in ```json ``` block in following JSON format:
{{
"answer": <either A or B>,
"explanation: "....",
}}
\end{Verbatim}
\end{tcolorbox}
Figure \ref{fig:knockout_prompt}: Knockout Prompt Template

\clearpage
\refstepcounter{boxcounter}
\begin{tcolorbox}[
  width=\textwidth,
  fontupper=\small\ttfamily,
  title={Knockout + Best-of-N Prompt $\phi_n$ (Yelp)}
]
\label{fig:best_of_n_prompt}
\begin{Verbatim}[breaklines=true,breaksymbol={}]
You are an impartial evaluator of style and content alignment. You will be evaluating a review text about a particular business written by an author. Below are samples of the author's writing containing past review texts of various businesses, review samples from other users for the business being reviewed, the input query, and three reviews.

# Author's Writing:
SAMPLE 1:
[SAMPLE 1 Text]
...

SAMPLE 2:
[SAMPLE 2 Text]
...

# Reviews from other users:
SAMPLE 1:
[SAMPLE 1 Text]
...

SAMPLE 2:
[SAMPLE 2 Text]
...

# Query:
[QUERY]

# Review A:
[GENERATED_OUTPUT_A]

# Review B:
[GENERATED_OUTPUT_B]

# Review C:
[GENERATED_OUTPUT_C]

# Task
Analyze the samples given under *Author's Writing* to identify the author's writing style. Consider the opinions expressed in *Reviews from other users* above to understand common themes and aspects of the business. Determine which of the three reviews is more likely to have been written by the author. Consider each review's similarity with regards to (1) tone (2) vocabulary level (3) sentence structure (4) tonal alignment with the target rating (5) avoidance of off-topic information (6) inclusion of relevant information based on other user reviews.

# Output Format:
You must return the winning review option along with a brief explanation for your choice.
Your output should be a valid json object in ```json ``` block in the following JSON format:
{{
"answer": <either A or B or C>,
"explanation: "....",
}}
\end{Verbatim}
\end{tcolorbox}
Figure \ref{fig:best_of_n_prompt}: Knockout + Best-of-N Prompt Template
\\
\\
The prompt templates for the other two datasets follow the same structure, with the word ‘business’ replaced by ‘book’ for Goodreads and by ‘product’ for Amazon throughout.
\clearpage
\refstepcounter{boxcounter}
\begin{tcolorbox}[
  width=\textwidth,
  fontupper=\small\ttfamily,
  title={Topic Extraction Prompts}
]
\label{fig:topic_extraction_prompts}
\tcbsubtitle{Writing Style Extraction $\phi_{ts}$}
\begin{Verbatim}[breaklines=true,breaksymbol={}]
You are a helpful assistant capable of analyzing a user's writing style. Your task is to evaluate a list of texts written by the user and determine the writing style. Consider elements such as tone, vocabulary, sentence structure, and overall flow.

# Your task
Given a list of writing samples by a user, analyze the writing style in terms of three aspects: tone, vocabulary, and sentence structure. If the samples have mixed sentiments (e.g., positive, negative, neutral), analyze and describe the writing style separately for each sentiment type. However, present your findings for each style aspect (tone, vocabulary, sentence structure) in a single, consolidated paragraph that includes distinctions based on sentiment. Avoid including specific content from the samples in your descriptions.

# Your inputs 
User's Writing : A list of sample texts written by the user.

# Your outputs - should be a valid JSON object in ```json``` block that contains the following keys. Limit the description to two or three sentences for each aspect:.
Tone : "..."
Vocabulary style: "..."
Sentence structure: "...
\end{Verbatim}
\tcbsubtitle{Content Topics Extraction $\phi_{tc}$}
\begin{Verbatim}[breaklines=true,breaksymbol={}]
You are a helpful assistant capable of extracting different atomic aspects discussed in a list of input texts. Your task is to analyze the texts and generate a JSON list of aspects and descriptions for each of them. Each aspect should be supported by evidence from the input texts, in the form of one or more related sentences.

# Your task
Given a list of review texts, identify common themes related to various characteristics of the reviewed item or experience, and extract atomic aspects. Additionally, generate a description for each identified aspect.

# Your inputs
Review texts : A list of review texts provided for a particular product, service, or entity.

# Your outputs - should be a valid JSON list in ```json``` block that contains the following keys.
aspect: the extracted aspect title
description: the description of the aspect.
\end{Verbatim}
\end{tcolorbox}
Figure \ref{fig:topic_extraction_prompts}: Topic Extraction Prompts.
\newpage
\refstepcounter{boxcounter}
\begin{tcolorbox}[
  width=\textwidth,
  fontupper=\small\ttfamily,
  title={Critic Prompt using extracted topics $\phi_{t}$ (Yelp)},
  breakable,
  enhanced
]
\label{fig:critic_topic_extraction}
\begin{Verbatim}[breaklines=true,breaksymbol={}]
You are a Personalization Critic Module. Specifically, you will analyze user review texts written for businesses. Your task is to evaluate how well the *Generated Output* is personalized for the target user and give feedback along the specified criteria.
        
# Your Inputs
Query - The user's request containing the business details.
Rating – the overall rating the target user has given for the product.
Generated Output - review text to be evaluated.
Writing style -  The writing style of the target user.
Business characteristics– A list of aspects and their descriptions about the business, based on reviews from other users.

# Feedback Criteria
- Tone Consistency: Does the tone and sentiment in the Generated Output align with the user’s writing style and the given review rating?
- Vocabulary Match: Does the vocabulary level in the Generated Output align with that of the user?
- Sentence Structure: Are the sentence lengths and structures similar to the user’s writing?
- Topic Relevance: Is the output content relevant to the query, free of off-topic information, and inclusive of important aspects?

Your feedback for the first three criteria (Tone Consistency, Vocabulary Match, Sentence Structure) should be based solely on the **Writing style** and should not include any content-related suggestions.
Your feedback for the last criterion (Topic Relevance) should be based solely on the **Business characteristics** input and should address the content.
               
# Output Format
Return your output in the following JSON format:
{
"Tone Consistency": ...,
"Vocabulary Match":  ...,
"Sentence Structure": ...,
"Topic Relevance": ...,
}

## Test Inputs

#Query:
[QUERY]

#Generated Output:
[GENERATED_OUTPUT]

#Writing style:

Tone: ...
Vocabulary style: ...
Sentence structure: ...

#Rating:
[TARGET_RATING]

#Business characteristics:

aspect title: ...
aspect detail: ...

aspect title: ...
aspect detail: ...

Your output should be a valid json object in ```json ``` block without any explanation.
\end{Verbatim}
\end{tcolorbox}
Figure \ref{fig:critic_topic_extraction}: Critic prompt using extracted topics.
\clearpage
\subsection{Evaluation Prompt}
\label{subsec:evaluation_prompts}
The prompt is borrowed from \cite{salemi2025experteffectiveexplainableevaluation}. G-Eval computes the final score by averaging over the possible scores, weighted by the probabilities assigned to each score by the backbone LLM. It was run 20 times at a temperature of 1 to obtain the probability distribution over scores.
\refstepcounter{boxcounter}
\begin{tcolorbox}[
  width=\textwidth,
  fontupper=\small\ttfamily,
  title={G-Eval (Pointwise scoring)}
]
\label{fig:geval}
\begin{Verbatim}[breaklines=true,breaksymbol={}]
You are a helpful assistant. Please act as an impartial judge and evaluate the quality of the response to instruction of the user displayed below. Based on the scoring criteria, please provide a score to the response compared to the reference. Be as objective as possible. You should consider both content and writing style similarity to assign a score.
    
# Your inputs:
-instruction: the instruction provided to the AI assistant.
-reference: the correct answer to the instruction.
-response: the response generated by the AI assistant.

# Scoring Criteria:
You should assign a score to the response based on the following criteria:
-Score 0: The answer is completely unrelated to the reference.
-Score 1: The answer has minor relevance but does not align with the reference.
-Score 2: The answer has moderate relevance but contains inaccuracies.
-Score 3: The answer aligns with the reference but has minor omissions.
-Score 4: The answer is completely accurate and aligns perfectly with the reference.

# Output:
Your output should be a valid JSON object in '''json''' block that contains the following keys:
-score: the score that you assigned to the AI assistant’s answer. The score should be an integer between 0 and 4.
\end{Verbatim}
\end{tcolorbox}
Figure \ref{fig:geval} : G-Eval Prompt.
\vspace{8pt}

\refstepcounter{boxcounter}
\begin{tcolorbox}[
  width=\textwidth,
  fontupper=\small\ttfamily,
  title={G-Eval-Style}
]
\label{fig:geval-style}
\begin{Verbatim}[breaklines=true,breaksymbol={}]
You are a helpful assistant. Please act as an impartial judge and evaluate the quality of the response to instruction of the user displayed below. Based on the scoring criteria, please provide a score to the response compared to the reference. Be as objective as possible. You should consider only the similarity in writing style, not the content, when assigning a score . Consider similarity in terms of tone, vocabulary, sentence structure, and overall flow.
    
# Your inputs:
-instruction: the instruction provided to the AI assistant.
-reference: the correct answer to the instruction.
-response: the response generated by the AI assistant.

# Scoring Criteria:
You should assign a score to the response based on the following criteria:
-Score 0: The writing style in the answer is completely different from the reference.
-Score 1: The writing style in the answer has minor similarities but does not align with the reference.
-Score 2: The writing style in the answer moderately matches the reference but not completely.
-Score 3: The writing style in the answer largely aligns with the reference.
-Score 4: The writing style in the answer is completely accurate and aligns perfectly with the reference.

# Output:
Your output should be a valid JSON object in '''json''' block that contains the following keys:
-score: the score that you assigned to the AI assistant’s answer. The score should be an integer between 0 and 4.
\end{Verbatim}
\end{tcolorbox}
Figure \ref{fig:geval-style} : G-Eval-Style Prompt.

\vspace{8pt}

\refstepcounter{boxcounter}
\begin{tcolorbox}[
  width=\textwidth,
  fontupper=\small\ttfamily,
  title={G-Eval-Content}
]
\label{fig:geval-content}
\begin{Verbatim}[breaklines=true,breaksymbol={}]
You are a helpful assistant. Please act as an impartial judge and evaluate the quality of the response to the instruction of the user displayed below. Based on the scoring criteria, please provide a score to the response compared to the reference. Be as objective as possible. You should consider only the similarity in terms of content, not the writing style, when assigning a score .
    
# Your inputs:
-instruction: the instruction provided to the AI assistant.
-reference: the correct answer to the instruction.
-response: the response generated by the AI assistant.

# Scoring Criteria:
You should assign a score to the response based on the following criteria:
-Score 0: The content in the response is completely different from the reference.
-Score 1: The content in the response has minor similarities but does not align with the reference.
-Score 2: The content in the response moderately matches the reference but not completely.
-Score 3: The content in the response largely aligns with the reference.
-Score 4: The content in the response is completely accurate and aligns perfectly with the reference.

# Output:
Your output should be a valid JSON object in '''json''' block that contains the following keys:
-score: the score that you assigned to the AI assistant’s answer. The score should be an integer between 0 and 4.
\end{Verbatim}
\end{tcolorbox}
Figure \ref{fig:geval-content} : G-Eval-Content Prompt.

\vspace{8pt}
\clearpage
\refstepcounter{boxcounter}
\begin{tcolorbox}[
  width=\textwidth,
  fontupper=\small\ttfamily,
  title={G-Eval-Profile}
]
\label{fig:geval-profile}
\begin{Verbatim}[breaklines=true,breaksymbol={}]
You are an impartial evaluator of style and content alignment of a response text. You will evaluate a review text about a particular item written by an author. Your task is to assess how well a given response aligns with the writing style of a target author and the content relevance to the input query.

# Your inputs:

- instruction: the input prompt or question
- response: a review text to evaluate
- author’s writing: past reviews written by the same author

Analyze the samples under the author’s writing to identify the author’s writing style. Determine how likely it is that the given response was written by the author. Check for

Stylistic Alignment:
- tone (formal, casual, enthusiastic, critical, etc.)
- vocabulary level and word choice
- sentence structure and phrasing patterns
- level of detail and writing style
- tonal alignment with the target rating in the query

Content Relevance:
- addresses the given query
- considers the content preferences of the author
- discusses relevant aspects of the item
- avoids off-topic or irrelevant information

# Scoring Rubric (0–4)

0 - No alignment
The response is inconsistent in both writing style and relevance to the query.
1 - Very weak alignment
Minimal similarity in style and/or poor relevance to the query.
2 - Moderate alignment
Some stylistic or content similarity, but clear inconsistencies remain.
3 - Strong alignment
The response mostly matches the author’s style and is relevant to the query, with minor issues.
4 - Very high alignment
The response closely matches the author’s writing style and is highly relevant to the query. It strongly resembles something the author would write.

# Output:
Your output should be a valid JSON object in '''json''' block that contains the following keys:
-score: the score that you assigned to the AI assistant’s answer. The score should be an integer between 0 and 4.
\end{Verbatim}
\end{tcolorbox}
Figure \ref{fig:geval-profile} : G-Eval-Profile Prompt.

\noindent
\clearpage
\subsection{Case study}
\label{subsec:example}
\begin{tcolorbox}[
  fontupper=\small\ttfamily,
  enhanced, breakable, width=1\textwidth,
  breakable
]
\tcbsubtitle{QUERY}
\begin{Verbatim}[breaklines=true,breaksymbol={}]
Generate a review text written by a user who has given an overall rating of 3.0 for a business with following details: 
 city: Tampa
 state: FL
 attributes:
 -RestaurantsGoodForGroups: True
 -GoodForKids: True
 -RestaurantsDelivery: False
 -RestaurantsAttire: 'casual'
 -RestaurantsReservations: True
 -BYOBCorkage: 'no'
 -BusinessAcceptsCreditCards: True
 -RestaurantsTakeOut: False
 -RestaurantsPriceRange2: 2
 -Alcohol: 'full_bar'
 -BusinessParking: 
    - garage : True
    - street : True
    - validated : False
    - lot : False
    - valet : True
 -OutdoorSeating: False
 categories: Restaurants, Asian Fusion
\end{Verbatim}
\tcbsubtitle{GROUNDTRUTH REFERENCE}
\begin{Verbatim}[breaklines=true,breaksymbol={}]
I really liked the interior. Besides being a food lover, I'm also a big fan of the ambiance. I had the lettuce wrap with chicken and shrimp, which was good. I also drank their mango mojito...tasty. My main course was the BC Spicy chicken and shrimp (their version of Kung Pao) which was also yummy. Our waitress/bartender was very nice and helpful. The food didn't blow my taste buds though, I guess I was just expecting more since it was such a nice restaurant. It is definitely worth a try and I will return for their happy hour.
\end{Verbatim}
\tcbsubtitle{USER PROFILE}
\begin{Verbatim}[breaklines=true,breaksymbol={}]
SAMPLE 1:
They use NY water! I really liked how they tried to capture NYC in their decorations as well. Its probably the best pizza in Tampa, but still not as good as you can get in NYC.  


I would really give them 3.5 stars.


SAMPLE 2:
Yummy! One of the best seafood restaurants in Tampa. The fish is so fresh and it tasted delicious = double points ( I was full, but could not stop eating.), the calamari was also great (You should try it because I have never had it served the way they made it.), presentation and service = GREAT. Everything was perfect from the moment we walked in. The ambiance, service, food...you get what you pay for and I seem to have had bad luck with that lately. I can't wait to come back and be a fatty again.


SAMPLE 3:
This place is awesome. Bowling, bars, music, and most importantly FOOD! =) I was also very impressed with their food selection; the variety they have can satisfy any person in your groups cravings/tastebuds. The tuna in their crunchy ahi poke dip was delicious and fresh. Oh, and they also have good drinks that can be served in huge glass bowls. A little pricey, but good for sharing. You have to go there.


SAMPLE 4:
I've been meaning to post this update sooner, but I was extremely busy. I really appreciated that they take customer satisfaction seriously. I was contacted by one of the owners to come back in for a complimentary meal, as a result of my previous review. The manager there told me that they made changes and was very adamant in having us try the new and/or revised dishes.  


From what we remembered last year (yes, it's been a year), the food has improved. We had their chips, salsa, guacamole, tacos, a fiesta bowl, and elote.


I still love their chips and salsa. Even though the chips have changed from last year, it was an improvement. They also now make homemade guacamole table-side, which was also very good. My friend usually hates guac, but decided to give it a try and loved it. The elote was good, although I wish it came out with char marks because in my opinion it tastes better that way. Service was also good. Our waitress checked up on us frequently.


Although it pains me to say what I'm going to say next, I can't help but be honest...I still was not impressed with their tacos or fiesta bowl. It was better than before, but not as good as I've had. For that reason, I would give them 3.5 stars.


I will still come here for drinks, chips, salsa and guacamole! If I was rating them just for those items and service, I'd give them 4 stars.
\end{Verbatim}
\tcbsubtitle{NEIGHBOR PROFILE}
\begin{Verbatim}[breaklines=true,breaksymbol={}]
SAMPLE 1:
Rating of 3.0 with the review: We came here on the recommendation of it being "good asian food". We went in despite the instinct to run from any restaurant that tries to be more of a "production" or "show" than a fine place to eat.


The waiter was, as expected, over-friendly to the point of extreme annoyance as he tried to tell us what was "great" on the menu and how every dish was his favorite. Yeah, sure.


The food was OK, and when biased for the rest of the food I had while Tampa, I have to admit it was above average. But not by much..


SAMPLE 2:
Rating of 1.0 with the review: I have been spoiled by authentic Chinese restaurants in Tampa, so me going to Bamboo Club is bound to be a disappointment from the start. Do they have anything that is not deep fried? It is at least in the category of Asian Fusion to let you know that it is anything but authentic and very similar to a take-out fried meat chunks covered in sweet sauce place, except with Cheesecake Factory prices. The establishment is beautiful and it has a full bar with delicious mixed drinks. But, don't come here expecting something unique or you will be severely disappointed. If this place does so well, then maybe TGI Fridays should add stir-fry to their menu and have the same taste and success..


SAMPLE 3:
Rating of 4.0 with the review: I came here for lunch with my step-sister before a day of shopping at the International Plaza. To keep it light we split an order of their Mongolian Beef lunch special and I have to say that it tasted extremely good. The atmosphere of the restaurant is very upscale and relaxing and the service was perfect as well. Although you are having lunch, they still give you a large enough amount of food to feed two people and it's really a great deal for people who want the great taste, but are not in a good position to pay their dinner prices. I'll be back for sure..


SAMPLE 4:
Rating of 3.0 with the review: I took my wife, mother, and children there for Mother's Day. Online they offered a kid's menu, but at the restaurant, not so much. They screwed up our reservation, which they confirmed online. The server came by only twice to check our drinks, and they forgot to place 2 of our 6 orders, so while 4 out of the six ate, 2 sat there. One of the two was a child. The food, when it arrived, was excellent. The ambiance was very cool. The service was sub-par. Being a former server, I am a very good tipper, usually 25% - 30 % of the bill. This time, let's just say...not so much. Would I go back? If someone else were picking up the tab.
\end{Verbatim}
\tcbsubtitle{PGraphRAG}
\begin{Verbatim}[breaklines=true,breaksymbol={}]
They have a full bar, which is great. I also like how they have a variety of food options. It's probably one of the better Asian restaurants in Tampa, but still not as good as you can get in NYC. I'd give it 3.0 stars.
\end{Verbatim}
\tcbsubtitle{\sysname{} (Knockout)}
\tcbsubtitle{Feedback 0}
\textbf{TONE CONSISTENCY}: The tone in the generated review is somewhat consistent with the target rating of 3.0, expressing a moderate level of satisfaction. However, the user's profile shows a \textcolor{blue}{more playful and enthusiastic tone} in positive reviews, and a more critical tone in negative reviews, which is not fully reflected in the generated output.\\
\textbf{VOCABULARY MATCH}: The vocabulary level in the generated review is generally appropriate, but it \textcolor{violet}{lacks the specific and sometimes colloquial language} used in the user’s profile, such as 'double points' or 'You have to go there.'\\
\textbf{SENTENCE STRUCTURE}: The sentence structure in the generated review is simple and straightforward, which is somewhat consistent with the user's profile, but the \textcolor{orange}{user tends to use more complex structures and varied lengths} in their reviews.\\
\textbf{TOPIC RELEVANCE}: The generated review covers relevant topics such as the full bar and food variety, but it misses important aspects mentioned in the neighbor profile, such as the \textcolor{lightgreen}{service quality} and the kid-friendly environment. Additionally, the review could have included more specific details about the \textcolor{lightgreen}{food quality and atmosphere}.
\tcbsubtitle{Generation 1}
\begin{Verbatim}[breaklines=true,breaksymbol={}]
They have a full bar, which is awesome! I loved how they have a variety of food options, it's like they really try to capture the Asian vibe in their menu. You have to go there for their Asian fusion food, it's probably one of the better options in Tampa, but still not as good as you can get in NYC. The service was good, our server checked up on us frequently and was super friendly. I also appreciated that it's kid-friendly, my little ones had a blast while we were there. Although the food was good, it wasn't mind-blowing, which is why I'd give it 3.0 stars.
\end{Verbatim}
\tcbsubtitle{Feedback 1}
\textbf{TONE CONSISTENCY}: The tone in the Generated Output is somewhat consistent with the user's writing style, but it \textcolor{blue}{lacks the casual} and slightly sarcastic tone seen in SAMPLE 1 and SAMPLE 4 of the User Profile.\\
\textbf{VOCABULARY MATCH}: The vocabulary used in the Generated Output is generally appropriate, but \textcolor{violet}{it is more formal and less colloquial} than what the user typically uses, as seen in SAMPLE 1 and SAMPLE 4.\\
\textbf{SENTENCE STRUCTURE}: The sentence structure in the Generated Output is relatively simple and straightforward, which is somewhat similar to the user's style, but the \textcolor{orange}{user tends to use more varied and complex sentence structures}, as seen in SAMPLE 4.\\
\textbf{TOPIC RELEVANCE}: The Generated Output covers relevant aspects such as the food, service, and kid-friendliness, aligning well with the query. However, it could \textcolor{lightgreen}{include more specific details about the atmosphere} or parking, which are mentioned in the Neighbor Profile.
\tcbsubtitle{Generation 2}
They've got a sweet setup going on with that full bar, I mean, who doesn't love a good drink selection? I dug the variety of food options, it's like they're trying to give you a taste of Asia in Tampa, but let's be real, it's still not as good as what you'd get in the city. Our server was on it, super friendly and checked in on us often, but the food, while good, wasn't blowing my mind, which is why I'm giving 'em 3.0 stars. I did appreciate that it's kid-friendly, my little ones had a blast, and the ambiance was pretty cool, I mean, it's got that upscale vibe going on. Parking's a bit of a pain, but, you know, it's a trade-off for good food, right?
\tcbsubtitle{Feedback 2}
\textbf{TONE CONSISTENCY}: The tone in the generated output is relatively casual and slightly critical, which aligns with the user's writing style in the User Profile, especially in Sample 1 and Sample 4 where a casual and slightly critical tone is present.\\
\textbf{VOCABULARY MATCH}: The vocabulary used in the generated output is simple and straightforward, which matches the user's vocabulary level in the User Profile.\\
\textbf{SENTENCE STRUCTURE}: The sentence structures in the generated output are relatively simple and varied, similar to the structures found in the User Profile, especially in Sample 1 and Sample 4.\\
\textbf{TOPIC RELEVANCE}: The generated output covers relevant aspects such as food quality, service, and ambiance, but it \textcolor{lightgreen}{does not mention} the specific attributes like the \textcolor{lightgreen}{kid-friendly environment and the full bar}, which are important aspects mentioned in the query and the Neighbor Profile.
\tcbsubtitle{Generation 3}
\textcolor{navyblue}{They've got a nice setup going on with that full bar, offering a good selection of drinks. The variety of food options is great, it's like they're trying to give you a taste of Asia in Tampa. I liked that it's kid-friendly, my little ones had a blast, and the ambiance is pretty cool, with that upscale vibe going on. However, the food, while good, wasn't blowing my mind, which is why I'm giving 'em 3.0 stars. One thing that's worth noting is that parking can be a bit of a pain, but the good food and friendly service make up for it}.
\tcbsubtitle{Feedback 3}
\textbf{TONE CONSISTENCY}: The tone in the generated review is somewhat consistent with the user's profile, but it is more casual and \textcolor{blue}{less detailed} compared to some of the user's reviews. The user tends to be more specific and detailed in their descriptions, whereas the generated review is more general and concise.\\
\textbf{VOCABULARY MATCH}: The vocabulary used in the generated review matches the user's profile well, with simple and straightforward language. The user often uses casual and descriptive words which align with the generated review's vocabulary.\\
\textbf{SENTENCE STRUCTURE}: The sentence structures in the generated review are relatively simple and straightforward, which aligns with the user's profile. However, \textcolor{orange}{the user's past reviews often contain more complex sentence structures and varied lengths}.\\
\textbf{TOPIC RELEVANCE}: The generated review covers important aspects of the restaurant, such as the food, drinks, ambiance, and service, which are relevant to the query. However, it lacks specific details about the Asian fusion cuisine and the kid-friendly aspects mentioned in the neighbor profile. Additionally, it does not mention the parking situation, which was a point of discussion in the neighbor profile.
\tcbsubtitle{Generation 4}
\textcolor{navyblue}{They've got a nice setup going on with that full bar, offering a good selection of drinks. The variety of food options is great, it's like they're trying to give you a taste of Asia in Tampa. I liked that it's kid-friendly, my little ones had a blast, and the ambiance is pretty cool, with that upscale vibe going on. However, the food, while good, wasn't blowing my mind, which is why I'm giving 'em 3.0 stars. One thing that's worth noting is that parking can be a bit of a pain, but the good food and friendly service make up for it}.
\tcbsubtitle{Feedback 4}
\textbf{TONE CONSISTENCY}: The tone in the generated output is somewhat consistent with the user's writing style, but it \textcolor{blue}{lacks the detailed and slightly critical tone} seen in the user's profile reviews. The user tends to provide more specific criticisms and less general statements.\\
\textbf{VOCABULARY MATCH}: The vocabulary level in the generated output matches the user's profile, using straightforward and conversational language.\\
\textbf{SENTENCE STRUCTURE}: The sentence structure in the generated output is simple and direct, similar to the user's profile, but the \textcolor{orange}{user tends to use slightly more complex sentences with clauses}.\\
\textbf{TOPIC RELEVANCE}: The generated output covers relevant topics such as food, drinks, ambiance, and service, but it misses some important aspects highlighted in the neighbor profile, such as the \textcolor{lightgreen}{over-friendly waiter} and the reservation issues.
\tcbsubtitle{Final Generation}
I went to this Asian fusion place in Tampa and was impressed by their full bar and extensive drink selection. The variety of food options is definitely a plus, as they try to give you a taste of Asia right here in Tampa, but I have to say that the waiter was a bit too over-friendly for my taste, trying to tell us what was great on the menu and making it feel like a production rather than a genuine dining experience. The food was good, but not exceptional, and the ambiance is nice with that upscale vibe, but the parking situation was a bit of a hassle. They also seemed to have issues with reservations, which was a bit of a letdown. Overall, it's a decent spot, but nothing that would make me go out of my way for. I'd give them 3.0 stars.
\tcbsubtitle{Final Feedback}
\textbf{TONE CONSISTENCY}: The tone in the generated output \textcolor{blue}{aligns well with the user's writing style}, expressing a balanced and slightly critical perspective. The sentiment matches the 3.0 rating.\\
\textbf{VOCABULARY MATCH}: The vocabulary level in the generated output \textcolor{violet}{matches the user's profile, using casual and straightforward language}.\\
\textbf{SENTENCE STRUCTURE}: The \textcolor{orange}{sentence lengths and complexity in the generated output are similar to those used by the user}, maintaining a consistent style.\\
\textbf{TOPIC RELEVANCE}: The \textcolor{lightgreen}{generated output covers relevant aspects such as the full bar, food options, waiter friendliness, ambiance, parking, and reservations}, aligning with the user's experience. However, it could include more specific details about the food quality and variety, as mentioned in the neighbor profile.
\end{tcolorbox}

\twocolumn
This case study illustrates the refinement process over five iterations. The PGraphRAG baseline does provide an appropriate response for a three-star rating. However, it could still be better personalized.

We can observe how the critic incrementally aligns the tone across iterations. In feedback 1, it suggests using a more enthusiastic and playful tone. Once this is addressed in the refined generations, it then notes in feedback 2 that the user prefers a more casual style, while in feedback 4 and feedback 5 it recommends incorporating more specific descriptions. After these aspects are integrated, the final feedback comments that the text is well aligned, balanced, and slightly critical (Highlighted in \textcolor{blue}{blue}).

A similar pattern occurs with sentence structure, where the critic consistently suggests making it more varied and complex (in line with the user's style) across all feedback iterations. This is finally corrected after five iterations, with the final feedback confirming that it is well aligned (Highlighted in \textcolor{orange}{orange}).

When it comes to vocabulary, the critic clearly recommends adopting a more colloquial style (in line with the user's style) in feedback 1 and feedback 2. Unlike tone and sentence structure, this aspect is resolved early, after two iterations, and in the remaining feedback the critic signals that it is well aligned and requires no further improvement (Highlighted in \textcolor{violet}{violet}).

The critic is able to analyze the user’s writing samples, infer the style, and compare it against the baseline generation.

On the content side, to ensure topic relevance, it provides useful suggestions regarding common aspects of restaurant reviews, such as service quality, ambiance, and friendly staff. These aspects are also present in the ground truth, leading to better topical relevance overall (Highlighted in \textcolor{lightgreen}{green}).

Another important observation is that generation 3 and generation 4 are identical. The critic judged the initial generation 4 as less personalized compared to generation 3, as a result of which the newly generated output was \textbf{knocked-out} and generation 3 was advanced as generation 4 (Highlighted in \textcolor{navyblue}{navyblue}).

\end{document}